\documentclass[sigconf]{acmart}

\AtBeginDocument{%
  \providecommand\BibTeX{{%
    \normalfont B\kern-0.5em{\scshape i\kern-0.25em b}\kern-0.8em\TeX}}}

\usepackage{amsmath}
\usepackage{amssymb}
\usepackage{bbold}
\newcommand{\myparagraph}[1]{\vspace{3pt}\noindent{\bf #1}}
\newcommand{\squeezeup}{\vspace{-2.5mm}}
\usepackage[ruled,linesnumbered]{algorithm2e}

\setcopyright{none}
\renewcommand\footnotetextcopyrightpermission[1]{} % removes footnote with conference information in first column
\begin{document}

%%
%% The "title" command has an optional parameter,
%% allowing the author to define a "short title" to be used in page headers.
%% SADAF: WE MIGHT RECONSIDER THE TITLE LATER
\title{Explaining with Counter Visual Attributes and Examples}

%%
%% The "author" command and its associated commands are used to define
%% the authors and their affiliations.
%% Of note is the shared affiliation of the first two authors, and the
%% "authornote" and "authornotemark" commands
%% used to denote shared contribution to the research.
\author{Sadaf Gulshad}
\email{s.gulshad@uva.nl}
\affiliation{
  \institution{UvA-Bosch Delta Lab, University of Amsterdam}
  \city{The Netherlands}
}

\author{Arnold Smeulders}
\email{a.w.m.smeulders@uva.nl}
\affiliation{%
  \institution{UvA-Bosch Delta Lab, University of Amsterdam}
  \city{The Netherlands}
}

%% By default, the full list of authors will be used in the page
%% headers. Often, this list is too long, and will overlap
%% other information printed in the page headers. This command allows
%% the author to define a more concise list
%% of authors' names for this purpose.
\renewcommand{\shortauthors}{Gulshad and Smeulders, et al.}

%% The abstract is a short summary of the work to be presented in the
%% article.
\begin{abstract}
In this paper, we aim to explain the decisions of neural networks by utilizing multimodal information. That is counter-intuitive attributes and counter visual examples which appear when perturbed samples are introduced. Different from previous work on interpreting decisions using saliency maps, text, or visual patches we propose to use attributes and counter-attributes, and examples and counter-examples as part of the visual explanations. When humans explain visual decisions they tend to do so by providing attributes and examples. Hence, inspired by the way of human explanations in this paper we provide attribute-based and example-based explanations. Moreover, humans also tend to explain their visual decisions by adding counter-attributes and counter-examples to explain what is \textit{not} seen. We introduce directed perturbations in the examples to observe which attribute values change when classifying the examples into the counter classes. This delivers intuitive counter-attributes and counter-examples. Our experiments with both coarse and fine-grained datasets show that attributes provide discriminating and human-understandable intuitive and counter-intuitive explanations.  
\end{abstract}

%% Keywords. The author(s) should pick words that accurately describe
%% the work being presented. Separate the keywords with commas.
\keywords{Explainability, classification, attributes, counter-intuitive attributes, perturbations, adversarial Examples}

\maketitle
%% This command processes the author and affiliation and title
%% information and builds the first part of the formatted document.

\section{Introduction}

\begin{figure}[t]
    %\begin{center}
        \includegraphics[width=\linewidth, trim=0 0 0 0, clip]{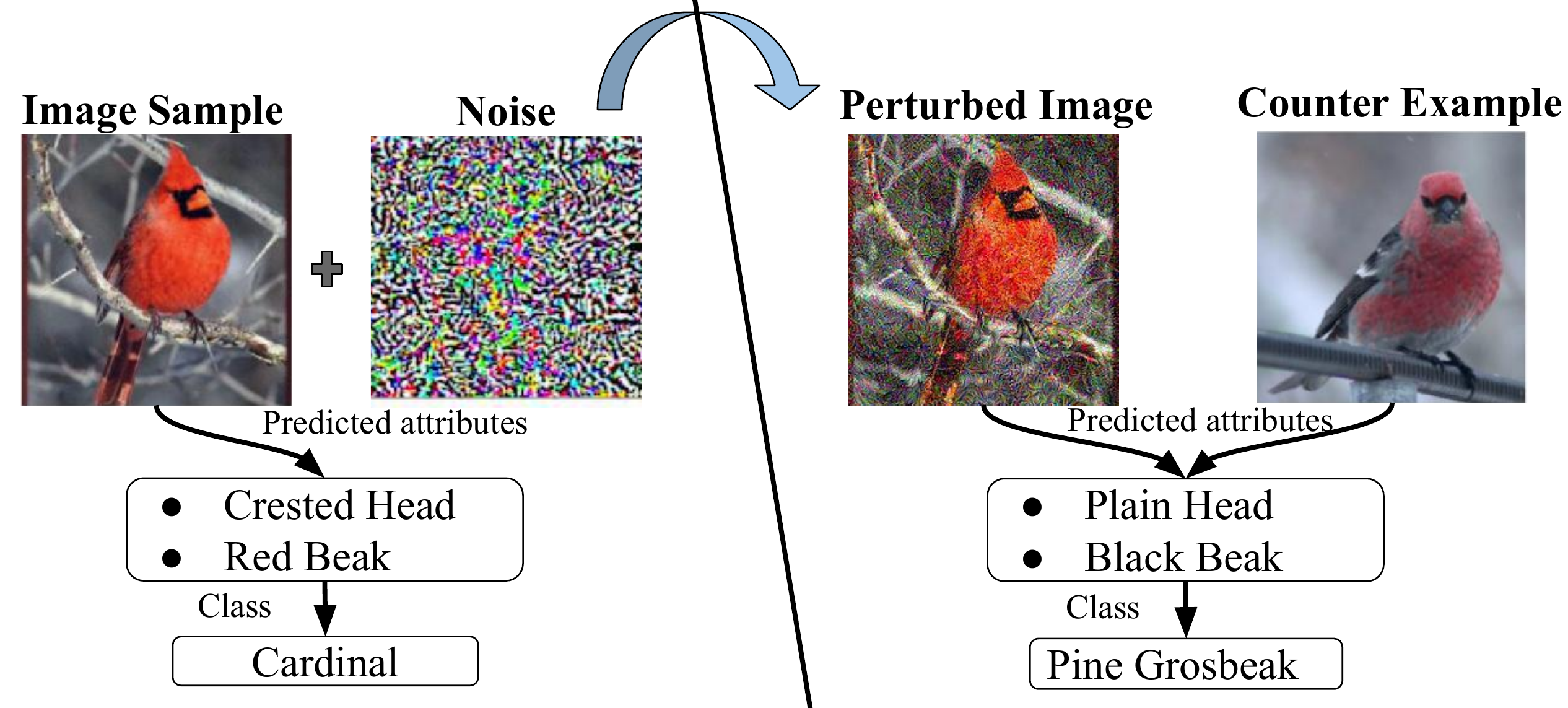}
    \caption{
     We use attributes to explain why an image on the left is classified into the \textit{Cardinal} class rather than the \textit{Pine Grosbeak} class on the right. And we use attributes with examples to explain when it will be classified as a \textit{Pine Grosbeak} by exploiting perturbed examples and their attribute values. We show that when the predicted attributes for the image change from ``\textit{Crested Head}'' and ``\textit{Red Beak}'' to ``\textit{Plain Head}'' and ``\textit{Black Beak}'', the image will be classified as \textit{Pine Grosbeak} .}
    \label{fig:Motivation}
    %\end{center}
    % \vspace{-3mm}
\end{figure}

When humans are asked to explain their visual decisions, we tend to do so by providing redundant, intuitive attributes visible in the image. For example, when a bird is classified as a \textit{``\textbf{Cardinal}''}, we might explain \textit{``because it has \textbf{crested head} and \textbf{red beak}''}. We tend to reinforce our explanations by adding what it is not with supporting attributes and examples. \textit{``When we would see a \textbf{plain head} and \textbf{black beak} rather than \textbf{crested head} and \textbf{red beak}, then the bird can no longer be a \textit{\textbf{Cardinal}} and should be classified as a \textit{\textbf{Pine Grosbeak}} ''}, see Figure~\ref{fig:Motivation}. Inspired by human explanations for decision making, we study attribute values under normal and perturbed conditions for the purpose of getting better explanations in this paper.

Understanding neural networks is crucial in applications like autonomous vehicles, health care, robotics, for validating and debugging, as well as for building the trust of users \cite{kim2018textual, uzunova2019interpretable}. This paper strives to explain the decisions of deep neural networks by studying the behavior of predicted attributes when an original image is classified into the correct class and when perturbed examples are introduced for classifying an image into the counter class ~\cite{szegedy2013intriguing}.

Most of the state of the art techniques for interpreting neural networks focus on saliency maps by considering class-specific gradient information \cite{selvaraju2017grad, simonyan2013deep, sundararajan2017axiomatic}, or by removing the part of the image which influences classification the most by adding perturbations \cite{zeiler2014visualizing, fong2017interpretable}. These approaches reveal where in the image there is support to the classification. However, the discrimination between the classes might reside in inconspicuous color or texture of the object or might be distributed over a large part of the image. Hence, it is difficult to explain discrimination by localisation. Therefore, these approaches tend to be weak in identifying the salient parts for fine-grained classification. While these approaches are valuable in their own right, they tell little about the contested important examples residing near the boundary of the class. 
In a recent work \cite{goyal2019counterfactual}, the authors proposed counter-factual explanations for both coarse and fine-grained classifications by replacing a region of the input image with the same region from an image of another class such that the class of the image changes. Our approach is steered with the same human-explanation motivation of using counter-examples. However, as their method works on the pixel level to edit the most discriminating region of an image, it requires a precise match of the imaging conditions (scale, viewpoint, scene and illumination) between the two images to work on. For most circumstances, this is difficult to achieve and is computationally expensive. In contrast, in this work, we propose to exploit perturbed examples to arrive at counter-attribute values as they change with a change in class. 

We complement our attribute-based explanations with examples containing these attributes. Example selection is related to the prototype selection\cite{singh2012unsupervised,doersch2012what}. The main idea of prototype selection is to extract the parts of data samples which represent all of the data \cite{molnar2019}. However, here we attempt to explain the decisions of neural networks by selecting the examples which demonstrate the critical attributes across the border of a class. Moreover, we utilize perturbations to find the counter classes and attributes for explanations, see Figure.~\ref{fig:Motivation}.

The main contributions of our work are as follows:

\begin{itemize}
    \item We provide novel explanations by selecting counter-attributes and counter-examples.
    \item We propose a novel method to select counter-examples based on counter-attributes predicted by introducing directed perturbations.
    \item We study the effect of change in attributes both \textit{quantitatively} and \textit{qualitatively}, for not only \textit{standard}, but also for \textit{robust} networks for which we propose a robustification measure.
    \end{itemize}
  
    Our results indicate that the method of explanation through attributes is effective on three benchmark data sets with varying size and granularity.

\section{Related Work}

Explaining the output of a decision maker is commonly motivated by the need to build user trust before deploying them into a real world environment.

\myparagraph{Explainability.}
Previous work for visual classification explanation is broadly grouped into two types: 1) \textit{rationalization}, that is, justifying the network's behavior and 2) \textit{introspective explanation}, that is, showing the causal relationship between input and the specific output \cite{du2018techniques}. The first group has the benefit of being human understandable, but it lacks a causal relationship between input and output. The second group incorporates the internal behavior of the network, but lacks human understandability. In this work, we explain the decisions of neural networks in the human style of explanations by singling out specific attributes for positive evidence when the image is classified correctly and by following specific attributes for negative evidence when the image is directed for misclassification in a counter class. 

An important group of work on understandibility focuses on text-based class discriminative explanations~\cite{hendricks2016generating,park2016attentive}, text-based interpretation with semantic information~\cite{dong2017improving} and generating counterfactual explanations with natural language~\cite{hendricks2018generating}, they all fall in the \textit{rationalization} category. Text-based explanations are orthogonal to our attribute-based explanations as attributes tend to deliver the key-words in the sentence and carry the quintessence for the semantic distinction. Especially for fine-grained classification all sentences for all classes might tend to display the same structure hence, the core of the semantic distinction between classes lie in attributes where we put our emphasis. Generating sentences is valuable but largely orthogonal to our approach.

\begin{figure}[t]
    \centering
        \includegraphics[width=\linewidth, trim=0 0 0 0, clip]{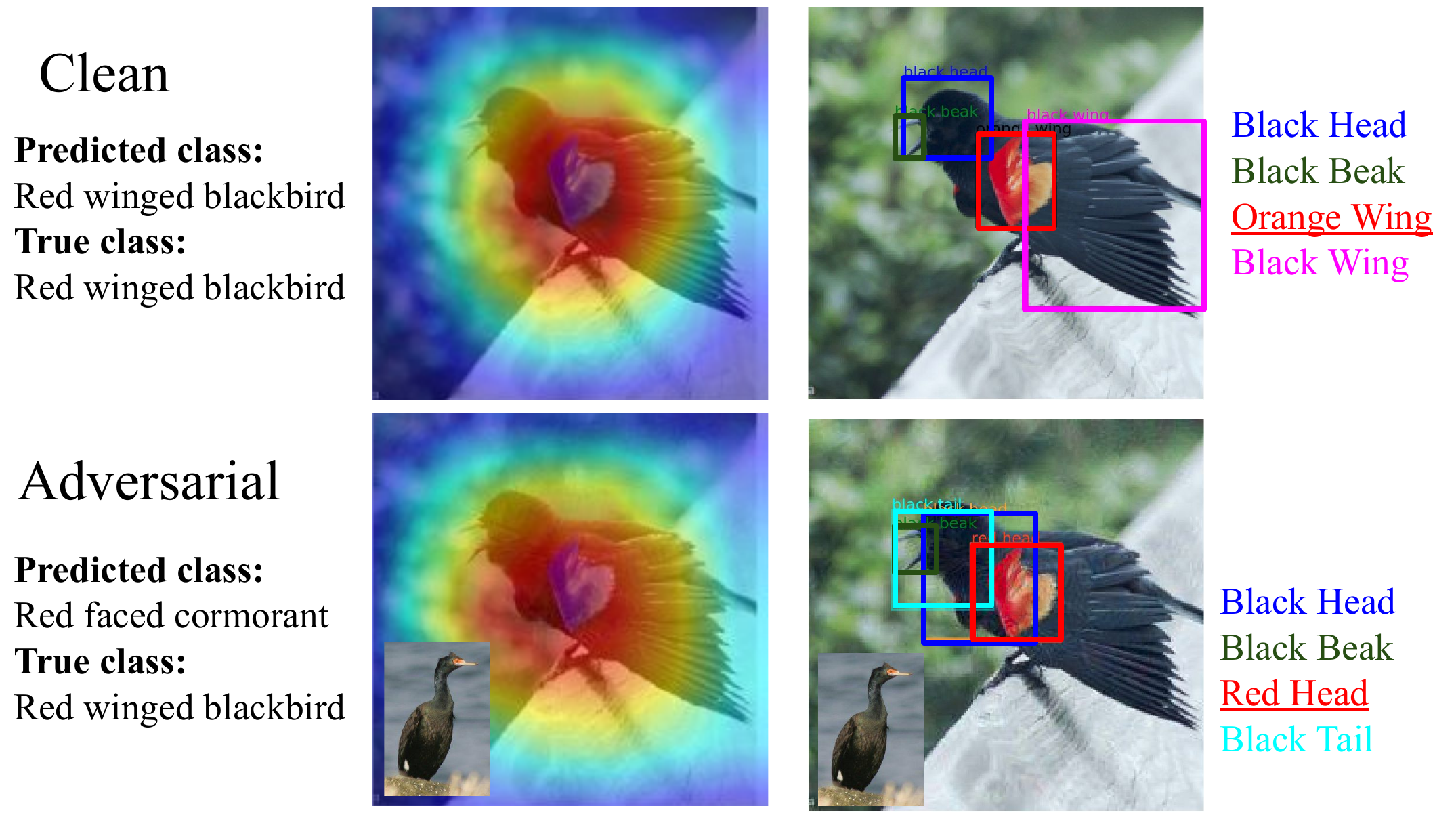}
    \caption{\textbf{Fine-grained images are difficult to explain with saliency maps:} when the answer is wrong, often saliency based methods (left) fail to detect what went wrong. Instead, attributes (right) provide intuitive and effective visual and textual explanations.}
  \squeezeup
  \vspace{-1mm}
    \label{fig:saliency}
\end{figure}

To tackle the similar task of explaining visual decisions, there is the large body of work on activation maximization \cite{simonyan2013deep, zintgraf2017visualizing}, learning the perturbation mask \cite{fong2017interpretable}, learning a model locally around its prediction, and finding important features by propagating activation differences \cite{ribeiro2016should,shrikumar2017learning}. They all fall in the group of \textit{introspective explanations}. All these approaches use saliency maps for explanation. We observe that saliency maps~\cite{selvaraju2017grad} are frequently weak in justifying classification decisions, especially for fine-grained images. For instance, in Figure~\ref{fig:saliency} the saliency map of a clean image classified into the ground truth class, ``red winged blackbird'', and the saliency map of a misclassified perturbed image, look quite similar. Instead, by grounding the predicted attributes, one may infer that the ``orange wing'' is important for ``red winged blackbird'' while the ``red head'' is important for ``red faced cormorant''. Indeed, when the attribute value for orange wing decreases and red head increases the image gets misclassified. Therefore, we propose to predict and ground attributes for both clean and perturbed images to provide visual as well as attribute-based interpretations. 

An interesting approach in a recent paper \cite{goyal2019counterfactual} proposes to generate counterfactual explanations by selecting a distractor image from a counter class and replacing the region in the input image with a region from the distractor image such that the class of the input image changes into the class of the distractor image. Pixel patch replacements pose high restrictions on the similarity of viewpoint, pose and scene between the two images, which makes the selection and replacement of the patches difficult. This is both computationally expensive. We follow the same inspiration of human-motivated counter examples. However, our approach focuses on attributes for generating explanations, as they contain the semantic core of the distinction between two competing classes and, attributes can naturally incorporate large changes in imaging conditions of size, illumination and viewpoint. Additionally, we use perturbations to change the class of the input image we analyze which attributes lead to the change in class. 

Another closely related work,  \cite{kanehira2019learning}, focuses on complementarity of multimodal information (i.e. text and examples) for explanations. They achieve this by maximizing the interaction information between the multimodal sources. However, by the nature of their method their example-based explanations will be visually completely different from the input image. In our work, by using the method of directed perturbations and discriminating attributes we are capable of selecting the most critical counter-examples as the most effective explanations.
% \vspace{-1.5mm}

\myparagraph{Adversarial Method} Small, carefully crafted perturbations, called \textit{adversarial perturbations}, have been used to alter the inputs of deep neural networks, which results in \textit{adversarial examples}. These adversarial examples drive the classifiers to the wrong class~\cite{szegedy2013intriguing}. Such methods of directed perturbations involve iterative fast gradient sign method (IFGSM) \cite{kurakin2016adversarial}, Jacobian-based saliency map attacks \cite{papernot2016limitations}, one pixel attacks \cite{su2019one}, Carlini and Wagner attacks \cite{carlini2017towards} and universal attacks \cite{moosavi2016deepfool}. Here, our aim is to utilize the directed noise from adversarial examples to study the change in attribute values therefore, we select the IFGSM-method for our experiments which leads the images into counter classes, because it is a fast and strong method.

\myparagraph{Adversarial Examples for Explainability.} Adversarial examples have been used for understanding neural networks. \cite{hsieh2020evaluations} aims at utilizing adversarial examples for understanding deep neural networks by extracting the features that provide the support for classification into the target class. After analyzing the neuronal activations of the networks for adversarial examples in \cite{dong2017towards}, it was concluded that networks learn recurrent discriminative parts of objects instead of semantic meaning. In \cite{jiang2018recent}, the authors proposed a data-path visualization module consisting of the layer level, the feature level, and the neuronal level visualizations of the network for clean as well as for adversarial images. All the above works focus on utilizing adversarial examples for understanding neural networks directly by analyzing feature maps or saliency maps. In contrast, we focus on exploiting adversarial examples to generate intuitive and counter-intuitive explanations with attributes and visual examples. 

In \cite{zhang2019interpreting}, the authors investigated adversarially trained robust convolutional neural networks by constructing input images with different textual transformations while at the same time preserving the shape information. They do this to verify the shape bias in adversarially trained networks compared with standard networks. Similarly, in \cite{tsipras2018robustness}, the authors showed that saliency maps from adversarially trained robust networks align well with human perception. 

In our work, we also provide explanations when an image is correctly classified with an adversarially trained robust network and verify that the attributes predicted by our method with a robust network still retain their discriminative power for explanations. 

\section{Method}

Given a clean $n\text{-th}$ input $x_n$ and its respective ground truth class $y_n$ predicted by a model $f(x_n)$, an adversarial model generates an image $\hat{x}_n$ for which the predicted class is $y$, where $y \neq y_n$.  In the following, we detail the adversarial method for perturbing a general classifier and an adversarial training technique that robustifies it. 

\subsection{Adversarial Methods}
    
\myparagraph{Adversarial Perturbation.} The iterative fast gradient sign method \cite{kurakin2016adversarial} (IFGSM) is leveraged to fool the classifier. IFGSM solves the following equation to produce adversarial examples:
\vspace{-1.5mm}
    \begin{align}
        & \hat{x}^0 =x_n \nonumber \\
        & \hat{x}_n^{i+1}=\text{Clip}_{\epsilon}\{\hat{x}_n^{i}+\alpha\text{Sign}(\bigtriangledown{\hat{x}_n^i}\mathcal{L}(\hat{x}_n^i,y_{n}))\}
    \end{align}
where $\bigtriangledown{\hat{x}_n^i}\mathcal{L}$ represents the gradient of the cost function w.r.t. perturbed image $\hat{x}_n^i$ at step $i$. $\alpha$ determines the step size which is taken in the direction of sign of the gradient and finally, the result is clipped by epsilon $\text{Clip}_{\epsilon}$.
    
\myparagraph{Adversarial Robustness.} We use \textit{adversarial training} as a robustness mechanism for the network when hardened by adversarial perturbation. This minimizes the following objective \cite{43405}:
    \begin{align}
        \mathcal{L}_{adv}(x_n,y_n) & = \alpha \mathcal{L}(x_n,y_n)
         + (1-\alpha)\mathcal{L}(\hat{x}_n,y)
    \end{align}
where, $\mathcal{L}(x_n,y_n)$ is the classification loss for clean images, $\mathcal{L}(\hat{x}_n,y)$ is the loss for adversarial images and $\alpha$ regulates the loss to be minimized. The model finds the worst case perturbations and fine tunes the network parameters to reduce the loss on perturbed inputs. Hence, this results in a robust network $f^r(\hat{x})$, using which improves the classification accuracy on the adversarial images.

 \begin{figure*}[t]
    \centering
    \includegraphics[width=\linewidth]{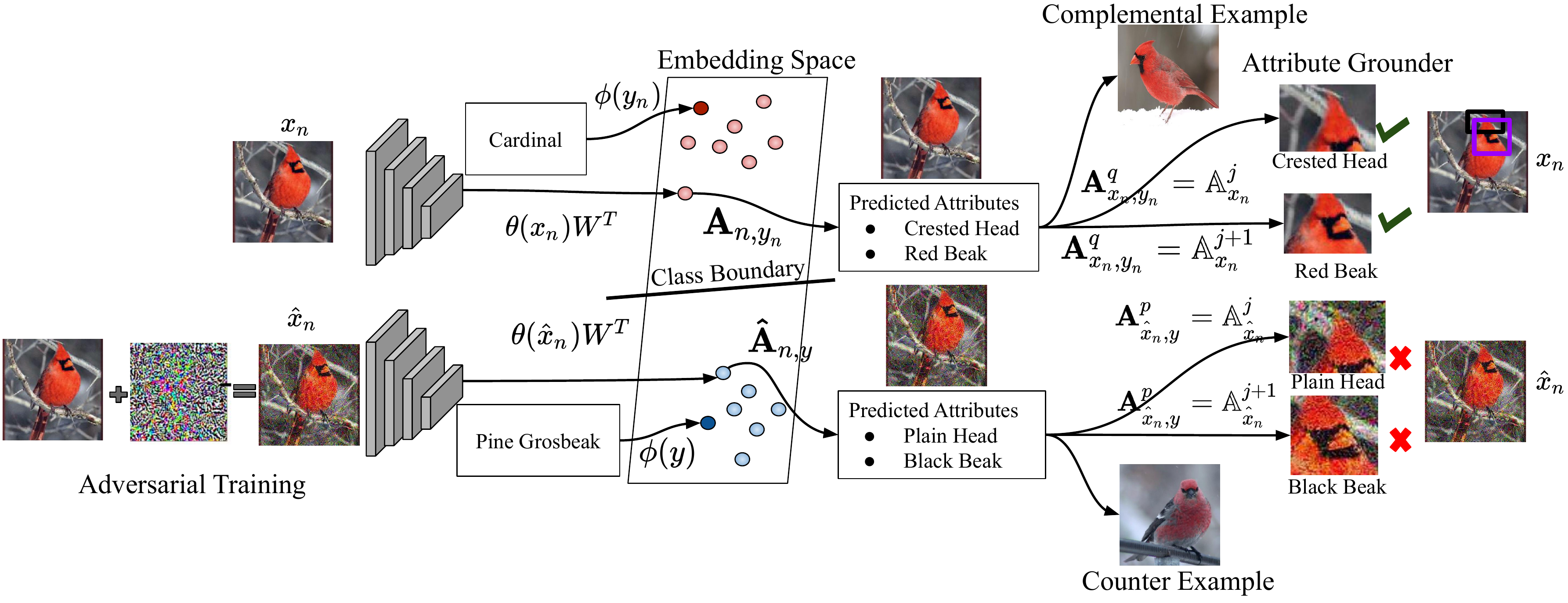}
    \squeezeup
    
    \caption{\textbf{Interpretable attribute prediction-grounding model.} After an adversarial training step, image features of both clean $\theta(x_n)$ and adversarial images $\theta(\hat{x})$ are extracted using Resnet and mapped into attribute space $\phi(y)$ by learning the compatibility function $F(x_n,y_n;W)$ between image features and class attributes. Finally, attributes predicted by attribute-based classifier $\bold{A}_{x_n,y_n}^q$ are grounded by matching them with attributes predicted by Faster RCNN $\mathbb{A}_{x_n}^j$ for clean and adversarial images. Examples are selected based on attribute similarity between adversarial image and adversarial class images for visual explanations.}
    \squeezeup
    \vspace{-1.5mm}
    \label{fig:ADV_SJE}
\end{figure*}  

\subsection{Attribute Prediction and Grounding}
Our attribute prediction and grounding model uses attributes to define a joint embedding space that the images are mapped to. 

\myparagraph{Attribute prediction.} The model is shown in Figure~\ref{fig:ADV_SJE}. During training our model maps clean training images close to their respective class attributes, and far from the attributes of other classes. At test time it maps clean images closer to their respective class attributes, e.g. ``Cardinal'' with attributes ``crested head, red beak'', whereas adversarial images get mapped close to a counter class, e.g. ``Pine Grosbeak'' with attributes ``plain head, black beak''. 

We employ structured joint embeddings (SJE)~\cite{akata2015evaluation} to predict attributes in an image. Given the input image features $\theta(x_n) \in \mathcal{X}$ and output class attributes $\phi(y_n) \in \mathcal{Y}$ from the sample set $\mathcal{S}=\{(\theta(x_n),\phi(y_n),n=1...N \}$  SJE learns a mapping $\mathbb{f}:\mathcal{X} \to \mathcal{Y}$ by minimizing the empirical risk of the form $\frac{1}{N}\sum_{n=1}^N \Delta(y_n,\mathbb{f}(x_n))$ where $\Delta: \mathcal{Y} \times \mathcal{Y} \to \mathbb{R} $ estimates the cost of predicting $\mathbb{f}(x_n)$ when the ground truth label is $y_n$.
    
A compatibility function $F:\mathcal{X}\times\mathcal{Y}\to \mathbb{R}$ is defined between input $\mathcal{X}$and output $\mathcal{Y}$ space:
\begin{equation}
        F(x_n,y_n;W)=\theta(x_n)^TW\phi(y_n)
\end{equation}
Pairwise ranking loss  $\mathbb{L}(x_n,y_n,y)$ is used to learn the parameters $(W)$:
    \begin{equation}
        \Delta(y_n,y)+\theta(x_n)^TW\phi(y_n)-\theta(x_n)^TW\phi(y)
    \end{equation}
Attributes are predicted for both clean and adversarial images by:
\vspace{-1.5mm}
\begin{equation}
    \bold{A}_{n,y_n}=\theta(x_n)W \, , \bold{\hat{A}}_{n,y}=\theta(\hat{x}_n)W 
\end{equation}
The image is assigned to the label of the nearest output class attributes $\phi(y_n)$.

\myparagraph{Attribute grounding.} Next, we ground the predicted attributes on to the input images using a pre-trained Faster RCNN network and visualize them as in~\cite{anne2018grounding}. The pre-trained Faster RCNN model $\mathcal{F}(x_n)$ predicts the bounding boxes denoted by $b^j$. For each object bounding box it predicts the class $\mathbb{Y}^j$ as well as the attribute $\mathbb{A}^j$ ~\cite{anderson2018bottom}.
\begin{equation}
    b^j,\mathbb{A}^j,\mathbb{Y}^j=\mathcal{F}(x_n)
\end{equation}
where $j$ is the bounding box index. 
We ground the most discriminative attributes predicted by SJE. They are selected based on the criterion that they change most when the image is perturbed with noise. For clean images we use: 
\begin{equation}
        q=\underset{i}{\mathrm{argmax}}(\bold{A}_{n,y_n}^i-\phi(y^i))
        \label{eq:att_sel1}
        \vspace{-1mm}
\end{equation}
and for adversarial images we use: 
\begin{equation}
        p=\underset{i}{\mathrm{argmax}}(\bold{\hat{A}}_{n,y}^i-\phi(y_n^i)). 
        \label{eq:att_sel2}
         \vspace{-1mm}
\end{equation}
where $i$ is the attribute index, and $q$ and $p$ are indexes of the most discriminative attributes predicted by SJE. $\phi(y^i)$, $\phi(y_n^i)$ 
indicate the wrong class and ground truth class attributes respectively. After selecting the most discriminative attributes predicted by SJE using equation \ref{eq:att_sel1} and \ref{eq:att_sel2} we match the selected attributes $\bold{A}_{x_n,y_n}^q, \bold{A}_{\hat{x}_n,y}^p$ with the attributes predicted by Faster RCNN for each bounding box $\mathbb{A}_{x_n}^j, \mathbb{A}_{\hat{x}_n}^j$. When the attributes predicted by SJE and Faster RCNN are matched, that is $\bold{A}_{x_n,y_n}^q = \mathbb{A}_{x_n}^j$, $\bold{A}_{\hat{x}_n,y}^p = \mathbb{A}_{\hat{x}_n}^j$ we ground them on their respective clean and adversarial images. 
\squeezeup

\subsection{Example Selection through Attributes.}
In order to provide example-based explanations we propose to select examples from the adversarial class through predicted attributes, the method is shown in Figure \ref{fig:ADV_SJE} and results in Figure \ref{fig:Qualitative-Examples-2}. The procedure for example selection from the adversarial class is given in Algorithm \ref{algo_sel}. Given adversarial and clean images and their predicted attributes our algorithm searches for images with the most similar attribute values in the adversarial class and selects them as counter examples. 
\IncMargin{1em}
\begin{algorithm}
\DontPrintSemicolon
\SetKwData{Left}{left}\SetKwData{This}{this}\SetKwData{Up}{up}
\SetKwFunction{Union}{Union}\SetKwFunction{FindCompress}{FindCompress}
\SetKwInOut{Input}{input}\SetKwInOut{Output}{output}

\Input{Adversarial images: $\hat{x}_{n,y}$, Clean images: ${x}_{n,y_n}$, Adversarial image attributes: $ \bold{\hat{A}}_{n,y}$, Clean image attributes: $ \bold{{A}}_{n,y_n}$, Adversarial classes: $y$}
\Output{Selected examples from adversarial class: $x^{s}_{n,y}$}
\BlankLine

\For{ each adversarial image $\hat{x}_{n,y}$}
{

Select all the images from adversarial class $x_{n,y}$ \\

\For{ each image in adversarial class $x_{n,y}$}
{$s=\underset{i}{\mathrm{argmin}}\parallel \bold{\hat{A}}^{{i}}_{n,y}-\bold{{A}}^{{i}}_{n,y} \parallel_2$
}
}\Return $x^{s}_{n,y}$
\caption{Example Selection through Attributes}\label{algo_sel}
\end{algorithm}\DecMargin{1em}

\squeezeup

\subsection{Attribute Analysis Method}
Finally, in this section we introduce our techniques for analysis on the predicted attributes.

\myparagraph{Predicted Attribute Analysis: Standard Network.} In order to perform analysis on attributes in embedding space, we consider the images which are correctly classified without perturbations and misclassified with perturbations.  Our aim is to analyse the change in attributes in embedding space.

We contrast the Euclidean distance between predicted attributes of clean and adversarial samples:
% %
\begin{equation}
    d_1 = d\{\bold{A}_{n,y_n},\bold{\hat{A}}_{n,y}\} =\parallel \bold{A}_{n,y_n}-\bold{\hat{A}}_{n,y} \parallel_2
    \label{eq:d1_1}
\end{equation}
with the Euclidean distance between the ground truth attribute vector of the correct and adversarial classes:
\begin{equation}
  d_2 = d\{\phi(y_n),\phi(y)\}=\parallel\phi(y_n)-\phi(y)) \parallel_2
  \label{eq:d2_1}
\end{equation}
 where, $\bold{A}_{n,y_n}$ denotes the predicted attributes for the clean images classified correctly, and $\bold{\hat{A}}_{n,y}$ denotes the predicted attributes for the adversarial images misclassified with a standard network. The correct ground truth class attribute is referred to as $\phi(y_n)$ and adversarial class attributes are $\phi(y)$.

\myparagraph{Predicted Attribute Analysis: Robust Network.}

We compare the distances between predicted attributes of only adversarial images that are classified correctly with the help of an adversarially robust network $\bold{\hat{A}}^{{r}}_{n,y_n}$ and classified incorrectly with a standard network $\bold{\hat{A}}_{n,y}$:
% % \squeezeup
\begin{equation}\label{eq:d1_3}
    d_1 = d\{\bold{\hat{A}}^{{r}}_{n,y_n},\bold{\hat{A}}_{n,y}\}=\parallel \bold{\hat{A}}^{{r}}_{n,y_n}-\bold{\hat{A}}_{n,y} \parallel_2
%     % \squeezeup
\end{equation}
with the distances between the ground truth target class attributes $\phi(y_n)$ and ground truth adversarial class attributes $\phi(y)$:
% % \squeezeup
\begin{equation}\label{eq:d2_3}
    d_2 = d\{\phi(y_n),\phi(y)\}=\parallel\phi(y_n)-\phi(y)) \parallel_2
\end{equation}
\subsection{Implementation Details}

\myparagraph{Image Features and Adversarial Examples.} We extract image features and generate adversarial images using the fine-tuned Resnet-152. Adversarial attacks are performed using the IFGSM-method with epsilon $\epsilon$ values $0.01$, $0.06$ and $0.12$. The $\l_\infty $ norm is used as a similarity measure between clean input and the generated adversarial example. 
 
\myparagraph{Adversarial Training.}
As for adversarial training, we repeatedly computed the adversarial examples while training the fine-tuned Resnet-152 to minimize the loss on these examples. We generated adversarial examples using the projected gradient descent method. This is a multi-step variant of FGSM with epsilon $\epsilon$ values $0.01$, $0.06$ and $0.12$ respectively for adversarial training as in~\cite{madry2017towards}.

\myparagraph{Attribute Prediction and Grounding.}
At test time the image features are projected onto the attribute space. The image is assigned with the label of the nearest ground truth attribute vector. The predicted attributes are grounded by using Faster-RCNN pre-trained on the Visual Genome Dataset since we do not have ground truth part bounding boxes for any of attribute datasets.

\section{Experiments and Results}
     \begin{figure*}[t]
        \centering
        \includegraphics[width=\linewidth, trim=0 0 0 0, clip]{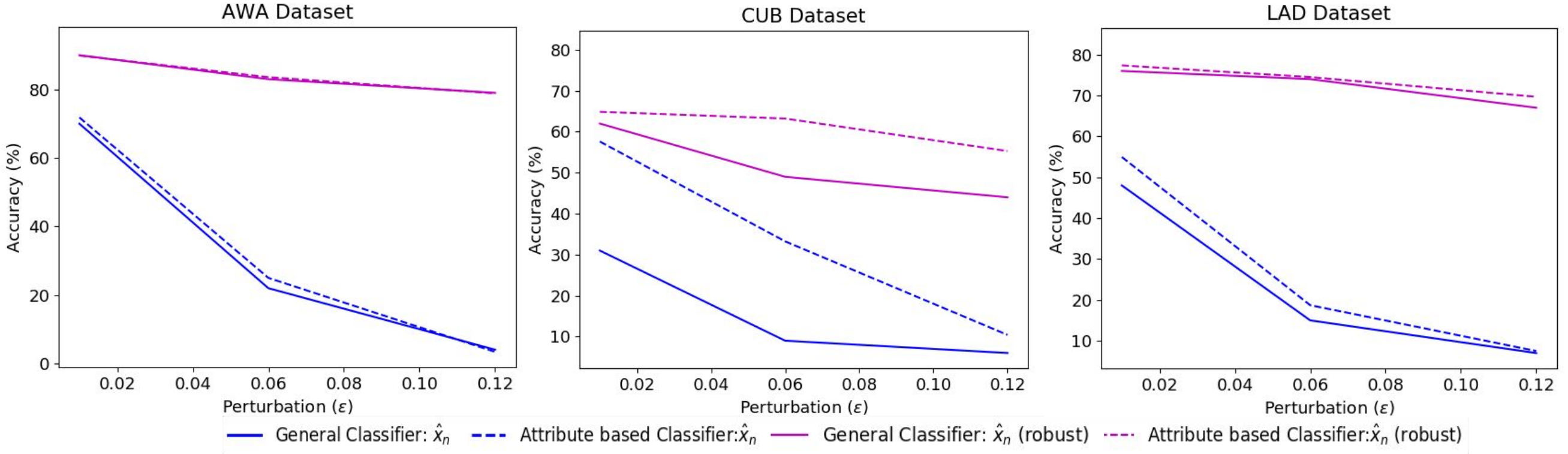}
        \squeezeup
        \squeezeup
        \caption{\textbf{Comparing the accuracy of the general and the attribute-based classifiers for adversarial examples to investigate change in attributes.} We evaluate both classifiers by extracting features from a standard network and the adversarially robust network.}
       \vspace{-1.5mm}
        \label{fig:acc_plots}
    \end{figure*}

\subsection{Datasets} 
We experiment on three datasets, Animals with Attributes 2 (AwA) \cite{lampert2009learning}, Large attribute (LAD) \cite{zhao2018large} and Caltech UCSD Birds (CUB)  \cite{wah2011caltech}. AwA contains 37322 images (22206 train / 5599 val / 9517 test) with 50 classes and 85 attributes per class. LAD has 78017 images (40957 train / 13653 val / 23407 test) with 230 classes and 359 attributes per class. CUB consists of 11,788 images (5395 train / 599 val / 5794 test) belonging to 200 fine-grained categories of birds with 312 attributes per class. All  three datasets contain real-valued class attributes representing the presence of an attribute in a class. For qualitative analysis with grounding we select $50$ attributes that change their value the most for the CUB, $50$ attributes for the AWA, and  $100$ attributes for the LAD dataset selected by equation \ref{eq:att_sel1} and \ref{eq:att_sel2}. 

The Visual Genome Dataset \cite{krishna2017visual} is used to train the Faster-RCNN model which extracts the bounding boxes using 1600 object and 400 attribute annotations. Each bounding box is associated with an attribute followed by the object, e.g. a brown bird. 

\subsection{Comparing General and Attribute-based Classifiers}
In the first experiment, we compare the general classifier $f(x_n)$ and the attribute-based classifier $\mathbb{f}(x_n)$ in terms of the classification accuracy on clean images to see whether the attribute-based classifier performs equally well.

We find that, the attribute-based and general classifier accuracies are comparable for AWA (general: 93.53, attribute-based: 93.83). The attribute-based classifier accuracy is slightly higher for LAD (general: 80.00, attribute-based: 82.77), and slightly lower for CUB (general: 81.00, attribute-based: 76.90) dataset. This indicates that both general and attribute-based classifiers perform equally well.
 \begin{figure}[t]
        \centering
        \includegraphics[width=\linewidth, trim=0 0 0 30, clip]{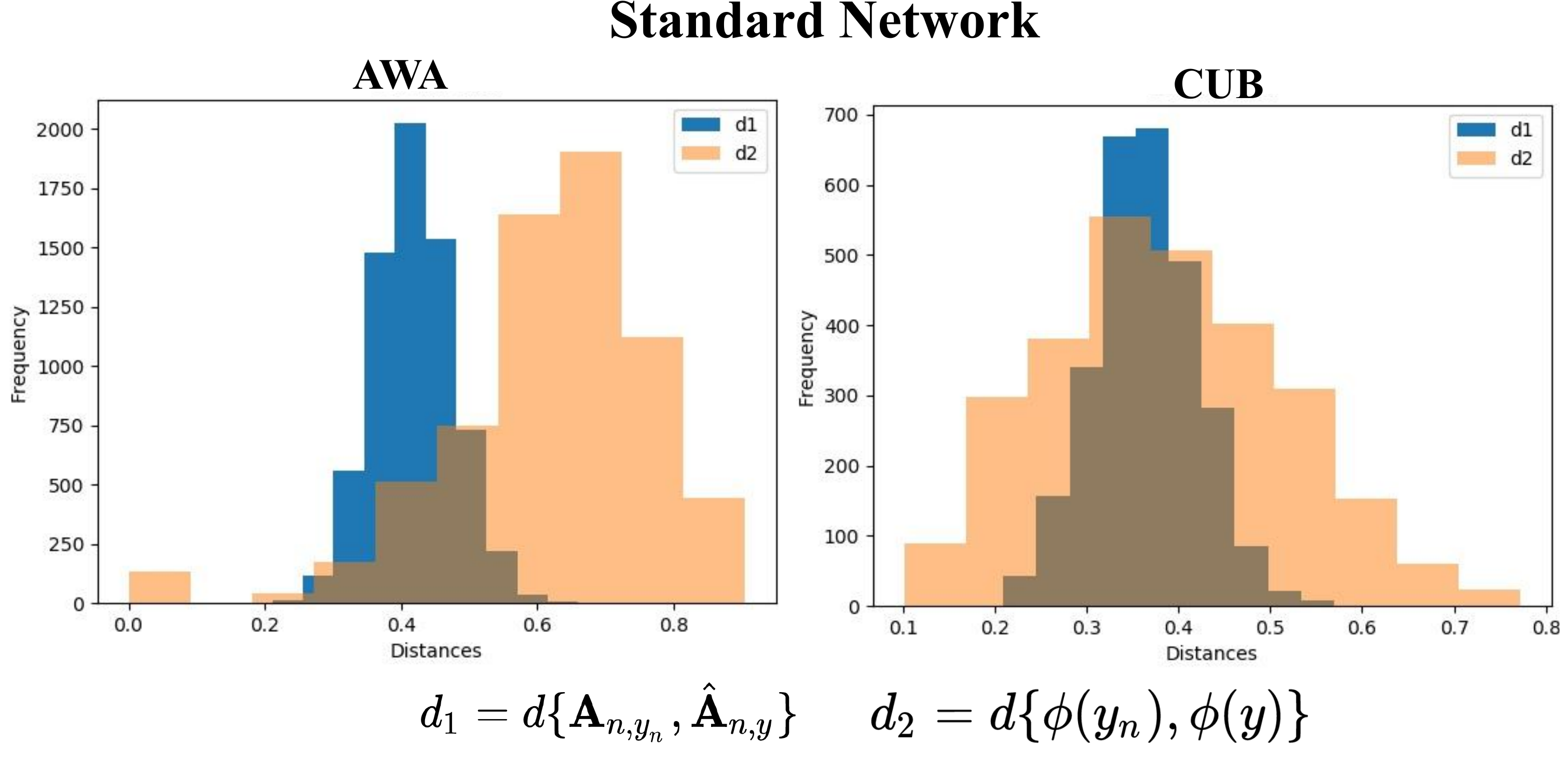}
        \vspace{-1.5mm}
        \vspace{-3mm}
        \caption{ Attribute value distance plots for clean and adversarial images with a standard network.}
            \label{fig:standardattr}
            \vspace{-3mm}
            \squeezeup
\end{figure}

\begin{figure}
    \centering
    \includegraphics[width=\linewidth, trim=0 0 0 0, clip]{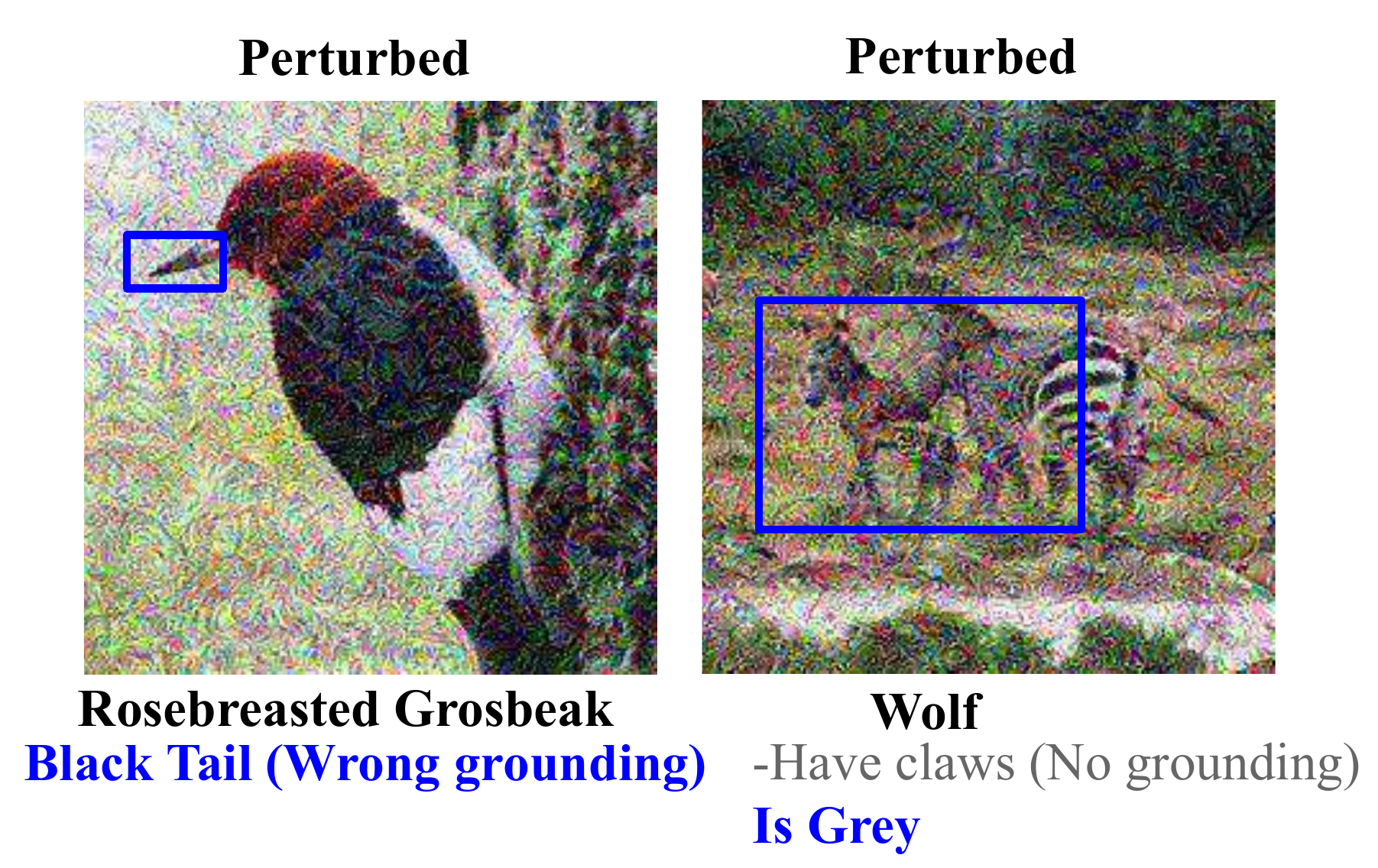}
    \vspace{-4.7mm}
    \caption{Explanation of a wrong classification due to wrong or missing attribute grounding. }
    \label{fig:Qualitative-1-1}
    \vspace{-3mm}
    \squeezeup
\end{figure}
\begin{figure*}[t]
    \centering
    \includegraphics[width=\linewidth, trim=0 5 0 0, clip]{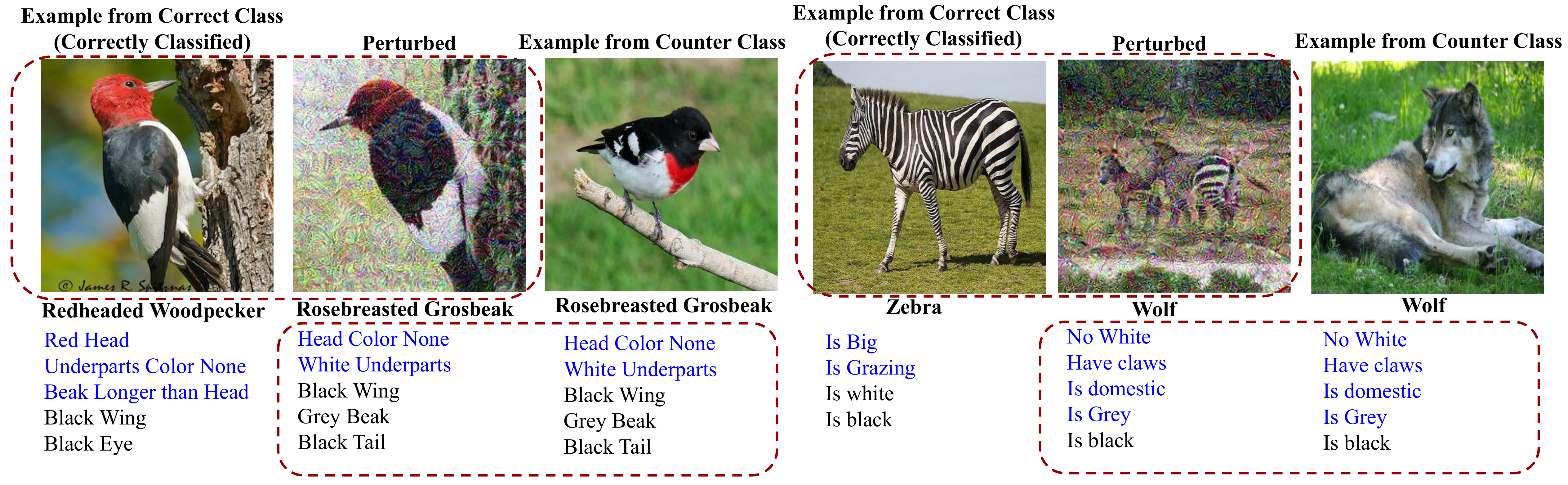}
    \vspace{-6mm}
    \caption{\textbf{Qualitative analysis for change in attributes due to directed perturbations with a standard network.} The attributes are ranked by importance for classification. }
    \label{fig:Qualitative-1}
    \vspace{-3mm}
\end{figure*} 
\subsection{Attribute-based Explanations: Standard Network}
In the second experiment we study the change in attributes with a standard network to demonstrate that by introducing perturbations in the images the attribute values change such that the class of the image changes to the counter-class and hence provide intuitive counter-explanations.
\subsubsection{By Performing Classification based on Attributes.} With adversarial attacks, the accuracy of both the general and attribute-based classifiers drops with the increase in perturbations see Figure~\ref{fig:acc_plots} (blue curves). The drop in accuracy of the general classifier for the fine-grained CUB dataset is higher as compared to the coarse-grained AWA dataset. For example, at $\epsilon=0.01$ for the CUB dataset the general classifier's accuracy drops from $81\%$ to $31\%$ ($\approx 50\%$ drop), while for the AWA dataset it drops from $93.53\%$ to $70.54\%$ ($\approx 20\%$ drop) and for LAD dataset it drops from $80.00\%$ to $50.00\%$ ($\approx 30\%$ drop). However, the drop in accuracy with the attribute-based classifier for CUB dataset is less $\approx 20\%$ as compared to general classifier. For coarse-grained datasets AWA and LAD the drop is almost the same for both classifiers. The smaller drop of accuracy for the CUB dataset in the attribute-based classifier when compared to the general classifier, is attributed to the fact that for fine-grained datasets there are many attributes common among classes. Therefore, in order to misclassify an image a significant number of attributes need to change their values. For a coarse-grained dataset, changing a few attributes is sufficient for misclassification. 
% \vspace{-0.5mm}

 Overall, the drop in the accuracy due to the pertubation demonstrates that the attribute values change towards those that belong to the new class. Hence attributes explain the misclassifications into the counter-classes well.

\subsubsection{By Computing Distances in the Embedding Space.} We contrast the Euclidean distance between predicted attributes of clean and adversarial samples using \ref{eq:d1_1} and \ref{eq:d2_1}. The results are shown in Figure~\ref{fig:standardattr}. We observe that for the AWA dataset the distances between the predicted attributes for adversarial and clean images $d_1$ are smaller than the distances between the ground truth attributes of the respective classes $d_2$. The closeness in predicted attributes for clean and adversarial images as compared to their ground truths shows that attributes change towards the wrong class but not completely. This is due to the fact that for coarse classes, only a small change in attribute values is sufficient to change the class.

The fine-grained CUB dataset behaves differently. The overlap between $d_1$ and $d_2$ distributions demonstrates that attributes of images belonging to fine-grained classes change significantly as compared to images from coarse categories. As the fine-grained classes are closer to one another so, many attributes are common among fine-grained classes, and require to change the attributes significantly to cause misclassification. Hence, for the coarse-grained dataset, the attributes change minimally, while for the fine-grained dataset they change significantly. 

\begin{figure*}[t]
    \centering
    \includegraphics[width=\linewidth, trim=0 0 0 0, clip]{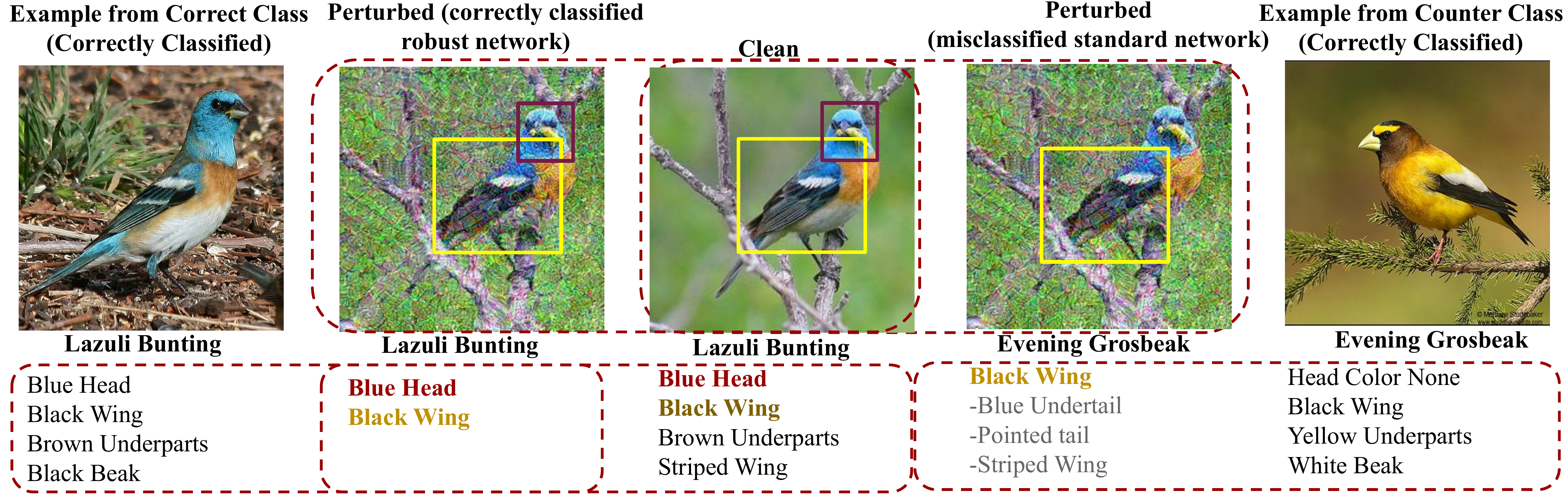}
    \vspace{-6mm}
    \caption{\textbf{Qualitative analysis for change in attributes due to directed perturbations with a robust network.} The attributes are ranked by importance for the classification decision, the grounded attributes are color coded for visibility (the ones in gray could not be grounded). }
    \label{fig:Qualitative-2}
    \vspace{-3mm}
\end{figure*} 

\subsubsection{Qualitative Analysis}

We observe in Figure~\ref{fig:Qualitative-1} that the most discriminative attributes for the clean images are coherent with the ground truth class however, for adversarial images they are coherent with the wrong class thus explaining the wrong class. For example ``red head, black wing, black eye'' attributes are responsible for the classification of clean image into correct class and when the value of ``red head'' attribute decreases and ``grey beak, white underparts'' increases the image gets misclassified into wrong class.

 Figure~\ref{fig:Qualitative-1-1} reveals the results for the groundings  on perturbed images. The attributes which are not related to the correct class, the ones that are related to the counter-class can not get grounded or get grounded at the wrong spots in the image as there is no visual evidence that supports the presence of these attributes. For example  ``black tail'' is related to the counter-class and is not present in the adversarial image. Hence,  black tail'' got wrongly grounded. This indicates that attributes for the clean images correspond to the ground truth class and for adversarial images correspond to the counter-class. Additionally, only those attributes common among both the counter and the ground truth classes get grounded on adversarial images.

Hence, our method provides explanations for both fine and coarse-grained classifications when the images get misclassified into similar classes or dissimilar classes. 

\squeezeup
\subsection{Attribute-based Explanations: Robust Network}
We perform the same experiments with a robust network to study the change in attribute values such that the class of the perturbed image changes back to the ground truth class.
\subsubsection{By Performing Classification based on Attributes.} Our evaluation on the standard and adversarially robust networks shows that the classification accuracy improves for the adversarial images when adversarial training is used to robustify the network Figure \ref{fig:acc_plots} (purple curves). For example, in Figure ~\ref{fig:acc_plots} for AWA the accuracy of the general classifier improved from $ 70.54\%$ to $92.15\%$ ($\approx 21\%$ improvement) and for LAD it improved from $ 50.00\%$ to $78.00\%$ ($\approx 28\%$ improvement) for adversarial attack with $\epsilon=0.01$. As expected for the fine-grained CUB dataset the improvement is $\approx 31\%$ higher than the AWA and LAD datasets. However, for the attribute-based classifier, the improvement in accuracy for AWA ($\approx 18.06\%$) is almost double and for LAD ($\approx 22.00\%$) almost triple that of the CUB dataset ($\approx 7\%$).  Which validates our results with the standard network that, for a fine-grained dataset the attributes need to change significantly to change the class. This further demonstrates that, attributes retain their discriminative power for explanations with standard as well as robust networks.

\subsubsection{By Computing Distances in the Embedding Space.}

With only adversarial images on robust and standard networks we observe the same distance distribution as Figure~\ref{fig:standardattr}. Thus, attributes explain the correct classification of adversarial images in the presence of the robust network.   
\vspace{-2.5mm}
\subsubsection{Qualitative Analysis}
% \hfill\\
Finally, our analysis with correctly classified images by the adversarially robust network shows that, adversarial images and their predicted attributes with the robust network behave like clean images and their predicted attributes as shown in Figure~\ref{fig:Qualitative-2}. This also demonstrates that, the attributes for adversarial images classified correctly with the robust network still retain their discriminative power and provide complementary explanations.
\begin{figure}[t]
    \centering
    \includegraphics[width=\linewidth, trim=0 0 0 0, clip]{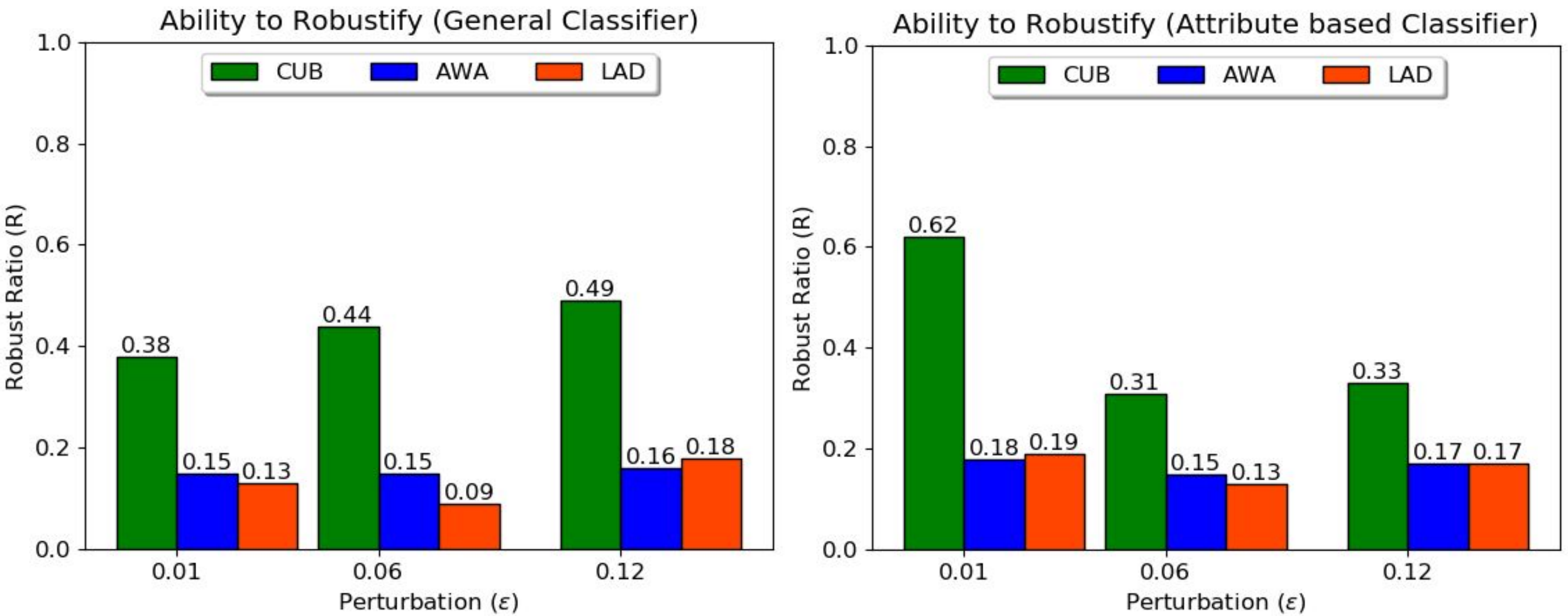}
    \vspace{-1mm}
    \caption{\textbf{Ability to robustify a network.} Ability to robustify a network with increasing adversarial perturbations is shown for three different datasets for both general and attribute-based classifiers.}
    \label{fig:Robustifiability}
    \squeezeup
    \squeezeup
\end{figure}
\begin{figure*}[t]
    \centering
    \includegraphics[width=\linewidth, trim=0 0 0 0, clip]{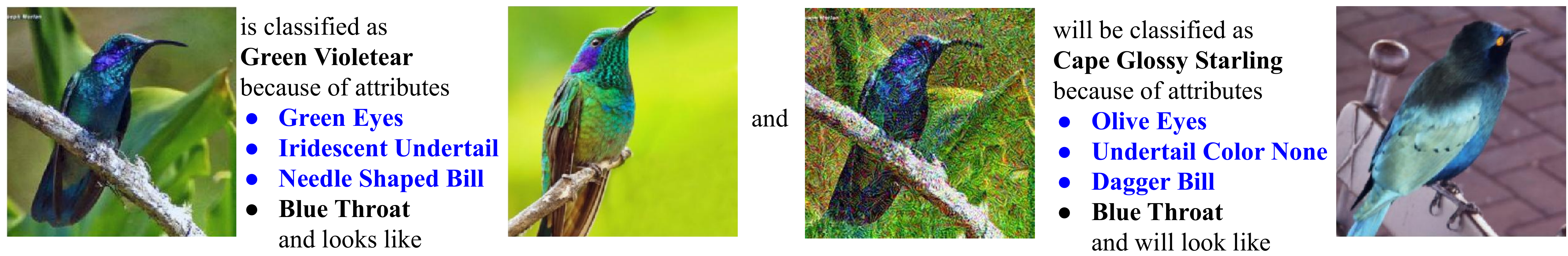}
    \vspace{-6mm}
    \caption{\textbf{Qualitative analysis for Example-based explanations. Note that when ``green eyes, needle shaped bill'' changes to ``olive eyes, dagger bill'' the class of the image changes.} }
    \label{fig:Qualitative-Examples-2}
    \vspace{-3mm}
\end{figure*}
\begin{figure*}[t]
    \centering
    \includegraphics[width=\linewidth, trim=0 0 0 0, clip]{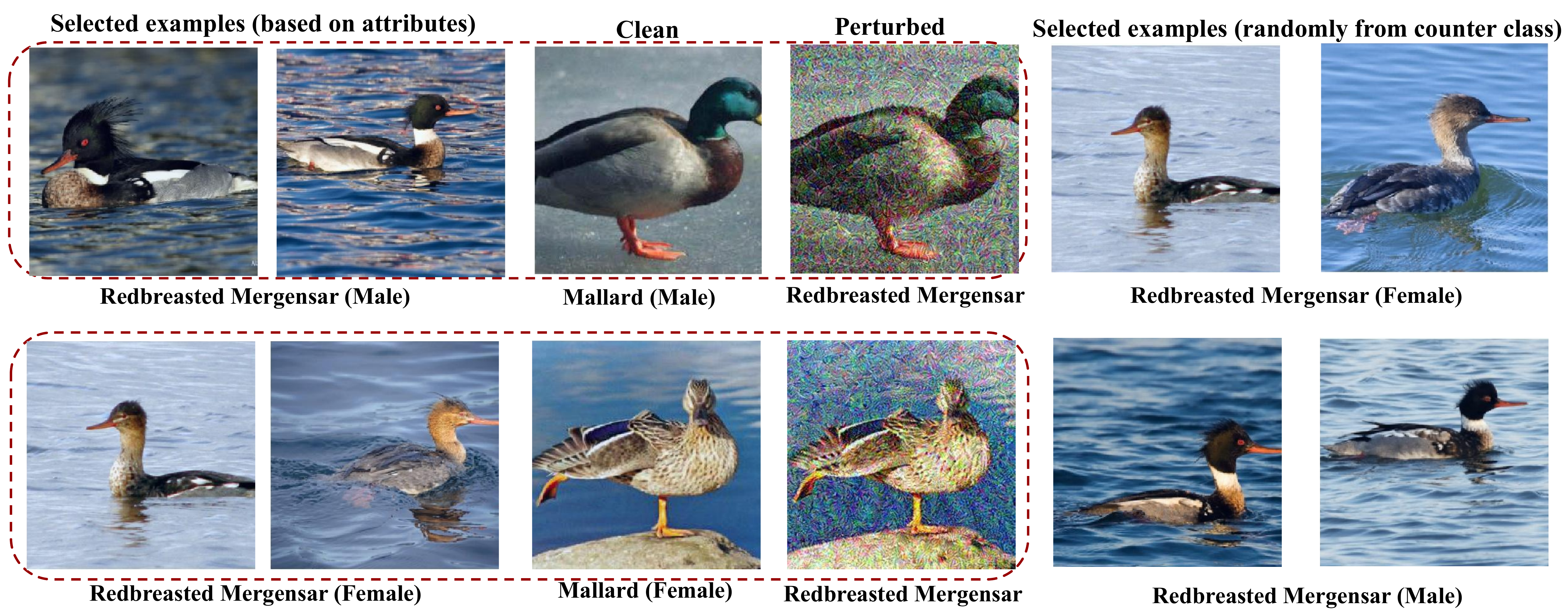}
    \vspace{-6mm}
    \caption{\textbf{Qualitative analysis for Example-based explanations. Note that both ``Mallard'' and ``Redbreasted Mergensar'' classes have inter-class variability as the male and female birds in both classes look visually different.} }
    \label{fig:Qualitative-Examples}
    \vspace{-3mm}
\end{figure*}

\subsubsection{By Robustness Quantification}
The results for our proposed robustness quantification metric are shown in Figure~\ref{fig:Robustifiability}. We observe that the ability to robustify a network against adversarial attacks varies for different datasets. The network with fine-grained CUB dataset is easy to robustify as compared to coarse-grained AWA and LAD datasets. For the general classifier as expected the ability to robustify the network increases with the increase in noise.

For the attribute-based classifier the ability to robustify the network is high with the small noise but it drops as the noise increases (at $\epsilon=0.06$) and then again increases at high noise value (at $\epsilon=0.12$).

\squeezeup
\subsection{ Example-based Explanations}
In experiment four we demonstrate our example and counter-example based explanations when the attribute values change with directed perturbations as shown in Figure \ref{fig:Qualitative-Examples-2}.  

Figure \ref{fig:Qualitative-Examples} reveals the importance of counter-example selection through attributes. In this example both the clean images in first and second row belong to same class ``Mallard'' however, the male and female Mallard differ visually. Similarly, the male and female birds of the counter-class ``Redbreasted Merganser'' also differ visually. The results for the examples retrieval for both male and female mallard show that, when images are retrieved through attributes for the male Mallard the retrieved images are male Redbreasted Merganser, while for the female Mallard the retrieved images through attributes are female Redbreasted Merganser. However, when we retrieve the images randomly from the counter-class then the visual similarity can not be ensured. Hence, our attribute-based example selection method selects the visually similar examples to provide distinction between clean image and counter-image classes especially when there is inter-class variation.

% \squeezeup

\section{Discussion and Conclusion}

In this work we conducted a systematic study on understanding neural networks by exploiting counter-examples with the attributes which causes misclassification. 

We first showed that attribute-based classifiers are as effective as direct classifiers. The efficacy of attributes for explanations is higher for the fine-grained datasets, where multiple attributes are shown to change their values between classes, as opposed to coarse-grained datasets, where only one attribute may suffice to change the class. 

In the second experiment, we demonstrated that we were able to explain decisions by selecting the most important counter-class for the given specific sample by moving to the closest counter-class by directed perturbation known from adversarial settings. We showed that if a noisy sample gets misclassified into a counter-class then its most discriminative attribute values indicate to which wrong class it is assigned. In the third experiment we verified that a noisy sample is correctly classified again with the robust version of the network: the discriminative attribute values return to the attribute values of the ground truth class, indicating the reliability of our directed perturbation. 
In the fourth experiment we have demonstrated the explanation of classification by showing a causal reason "because it contains this attribute and it does not contain that attribute." We also provide explanations by selecting the sharpest counter-example from the counter-class for the most relevant attribution. Our method gives the explanations by examples and counter-examples with the precise and most illustrative examples in the dataset. 

Hence we conclude that, attributes provide discriminative, human understandable, intuitive as well as counter-intuitive explanations for the neural networks especially for fine-grained classification. Even for the robust networks attributes retain their discriminative power and provide intuitive explanations.

\bibliographystyle{ACM-Reference-Format}
\bibliography{sample-base}
\end{document}